\documentclass[10pt,twocolumn,letterpaper]{article}

\usepackage{cvpr}
\usepackage{times}
\usepackage{epsfig}
\usepackage{graphicx}
\usepackage{amsmath}
\usepackage{amssymb}

\usepackage{graphicx}  
\usepackage{times}
\usepackage{epsfig}
\usepackage{graphicx}
\usepackage{amsmath}
\usepackage{amssymb}
\usepackage{ctable}
\usepackage{bm}
\usepackage{array,xcolor} 
\usepackage{tabularx} 
\newcolumntype{Y}{>{\centering\arraybackslash}X}
\newcolumntype{C}{>{\centering\arraybackslash}p{8ex}}
\newcolumntype{T}{>{\centering\arraybackslash}p{3.5ex}}

\usepackage{amsthm}
\usepackage{epstopdf}
\usepackage{amsthm}
\usepackage{multirow}
\usepackage{float}
\usepackage{enumitem}
\usepackage{graphicx}
\usepackage{subcaption}
\usepackage[pagebackref=true,breaklinks=true,letterpaper=true,colorlinks,bookmarks=false]{hyperref}

\cvprfinalcopy 

\def\cvprPaperID{2821} 

\ifcvprfinal\fi
\begin{document}

\title{Translating and Segmenting Multimodal Medical Volumes with Cycle- and Shape-Consistency Generative Adversarial Network}

\author{Zizhao Zhang$^{1,2}$, ~Lin Yang$^1$, ~Yefeng Zheng$^2$ \\
	$^1$University of Florida \\
	 $^2$Medical Imaging Technologies, Siemens Healthcare \\
}

\maketitle

\maketitle
\begin{abstract}
Synthesized medical images have several important applications, e.g., as an intermedium in cross-modality image registration and as supplementary training samples to boost the generalization  capability of a classifier. Especially, synthesized computed tomography (CT) data can provide X-ray attenuation map for radiation therapy planning. In this work, we propose a generic cross-modality synthesis approach with the following targets: 1) synthesizing realistic looking 3D images using unpaired training data, 2) ensuring consistent anatomical structures, which could be changed by geometric distortion in cross-modality synthesis and 3) improving volume segmentation by using synthetic data for modalities with limited training samples. We show that these goals can be achieved with an end-to-end 3D convolutional neural network (CNN) composed of mutually-beneficial generators and segmentors for image synthesis and segmentation tasks. The generators are trained with an adversarial loss, a cycle-consistency loss, and also a shape-consistency loss,  which is supervised by segmentors, to reduce the geometric distortion. From the segmentation view, the segmentors are boosted by synthetic data from generators in an online manner. Generators and segmentors prompt each other alternatively in an end-to-end training fashion. With extensive experiments on a dataset including a total of 4,496 CT and magnetic resonance imaging (MRI) cardiovascular volumes, we show both tasks are beneficial to each other and coupling these two tasks results in better performance than solving them exclusively.
\end{abstract}

\vspace{-.2cm}
\section{Introduction}
In current clinical practice, multiple imaging modalities may be available for disease diagnosis and surgical planning \cite{cao2017dual}.
For a specific patient group, a certain imaging modality might be more popular than others.
Due to the proliferation of multiple imaging modalities, there is a strong clinical need to develop a cross-modality image transfer analysis system to assist clinical treatment, such as radiation therapy planning \cite{burgos2015robust}. 

\begin{figure}[t]
	\begin{center}
		\includegraphics[width=0.49\textwidth]{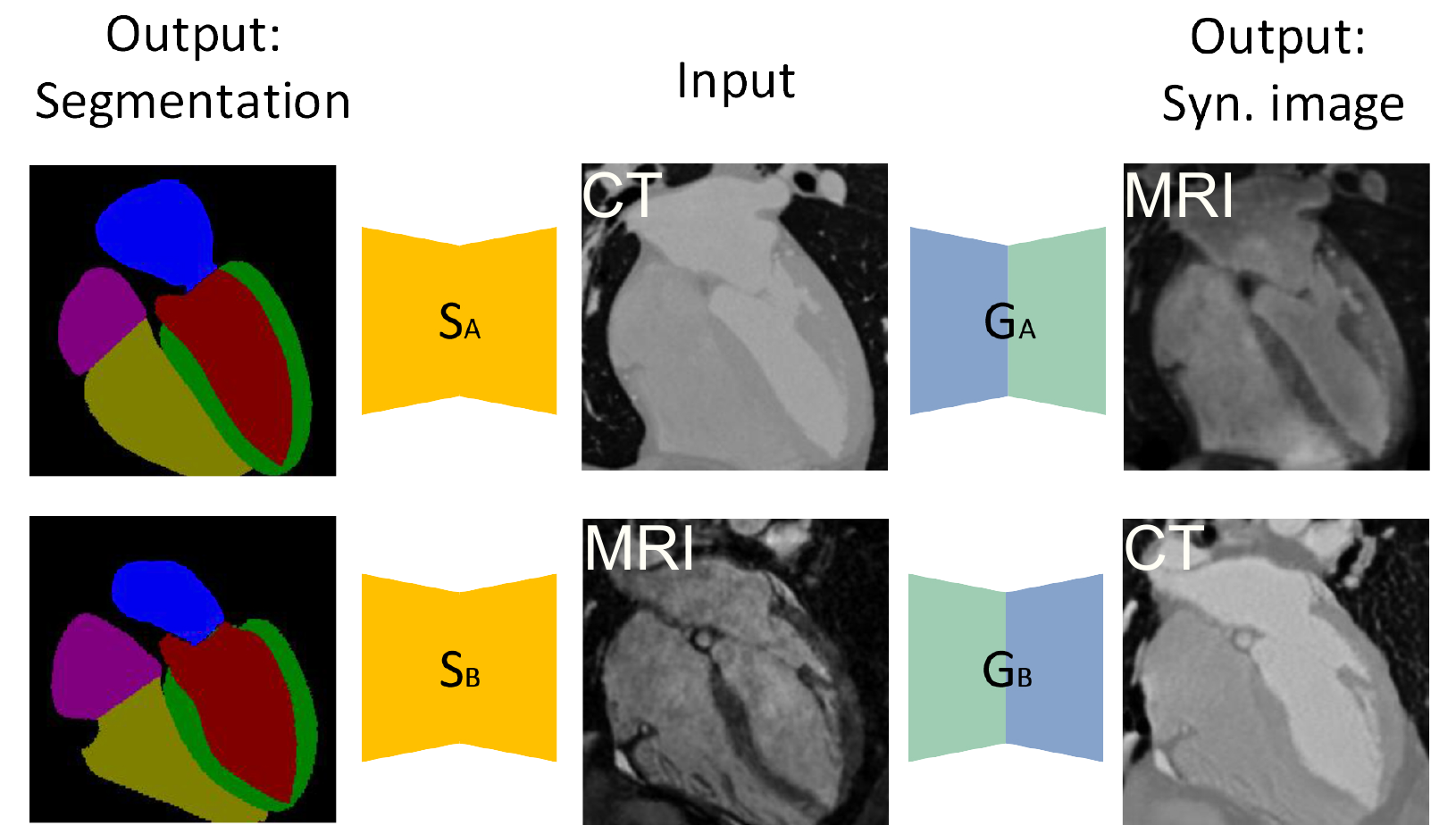}
	\end{center}
	\vspace{-.4cm}
	\caption{Our method learns two parallel sets of generators $G_{A/B}$  and segmentors $S_{A/B}$ for two modalities A and B to translate and segment holistic 3D volumes. Here we illustrate using CT and MRI cardiovascular 3D images.} \label{fig:intro} \vspace{-.2cm}
\end{figure}

Machine learning (ML) based methods have been widely used for medical image analysis \cite{Zhang_2017_CVPR,Zhang2017TandemNet}, including detection, segmentation, and tracking of an anatomical structure.
Such methods are often generic and can be extended from one imaging modality to the other by re-training on the target imaging modality.
However, a sufficient number of representative training images are required to achieve enough robustness.
In practice, it is often difficult to collect enough training images, especially for a new imaging modality not well established in clinical practice yet. 
Synthesized data are often used to as supplementary training data in hope that they can boost the generalization capability of a trained ML model.
This paper presents a novel method to address the above-mentioned two demanding tasks (Figure \ref{fig:intro}). The first is cross-modality translation and the second is improving segmentation models by making use of synthesized data.

To synthesize medical images, recent advances \cite{nie2016medical,costa2017towards} have used generative adversarial networks (GANs) \cite{goodfellow2014generative} to formulate it as an image-to-image translation task. These methods require pixel-to-pixel correspondence between two domain data to build direct cross-modality reconstruction. 
However, in a more common scenario, multimodal medical images are in 3D and do not have cross-modality paired data.
A method to learn from unpaired data is more general purpose.
Furthermore, tomography structures (e.g. shape), in medical images/volumes, contain diagnostic information. Keeping their invariance in translation is critical. However, when using GANs without paired data, due to the lack of direct reconstruction,
relying on discriminators to guarantee this requirement is not enough as we explain later.

It is an active research area by using synthetic data to overcome the insufficiency of labeled data in CNN training. 
In the medical image domain, people are interested in learning unsupervised translation between different modalities \cite{kamnitsas2017unsupervised}, so as to transfer existing labeled data from other modalities. 
However, the effectiveness of synthetic data heavily depends on the distribution gap between real and synthetic data.
A possible solution to reduce such gap is by matching their distributions through GANs \cite{shrivastava2016learning,bousmalis2016unsupervised}.

In this paper, we present a general-purpose method to realize both medical volume translation as well as segmentation. In brief, given two sets of unpaired data in two modalities, we simultaneously learn generators for cross-domain volume-to-volume translation and stronger segmentors by taking advantage of synthetic data translated from another domain.
Our method is composed of several 3D CNNs. 
From the generator learning view, we propose to train adversarial networks with cycle-consistency \cite{zhu2017unpaired} to solve the problem of data without correspondence. 
We then propose a novel shape-consistency scheme to guarantee the shape invariance of synthetic images, which is supported by another CNN, namely segmentor. From the segmentor learning view, segmentors directly take advantage of generators by using synthetic data to boost the segmentation performance in an online fashion. Both generator and segmentor can take benefits from another in our end-to-end training fashion with one joint optimization objective.

On a dataset with 4,496 cardiovascular 3D image in MRI and CT modalities, we conduct extensive experiments to demonstrate the effectiveness of our method qualitatively and quantitatively from both generator and segmentor views with our proposed auxiliary evaluation metrics.
We show that using synthetic data as an isolated offline data augmentation process underperforms our end-to-end online approach. 
On the volume segmentation task, blindly using synthetic data with a small number of real data can even distract the optimization when trained in the offline fashion. However, our method does not have this problem and leads to consistent improvement.

\section{Related work}
There are two demanding goals in medical image synthesis. The first is synthesizing realistic cross-modality images \cite{huang2017simultaneous,nie2016medical}, and second is to use synthetic data from other modalities with sufficient labeled data to help classification tasks (e.g. domain adaption \cite{kamnitsas2017unsupervised}).

In computer vision, recent image-to-image translation is formulated as a pixel-to-pixel mapping using encoder-decoder CNNs  \cite{isola2016image,liu2017unsupervised,zhu2017unpaired,kim2017learning,liu2017unsupervised,xue2017differential,gong2017learning}. 
Several studies have explored cross-modality translation for medical images, using sparse coding \cite{huang2017simultaneous,vemulapalli2015unsupervised}, GANs \cite{nie2016medical,osokin2017gans}, CNN \cite{van2015cross}, etc. GANs have attracted wide interests in helping addressing such tasks to generate high-quality, less blurry results \cite{goodfellow2014generative,arjovsky2017wasserstein,berthelot2017began,zhang2018txt2img}.
More recent studies apply pixel-to-pixel GANs for brain MRI to CT image translation \cite{nie2016medical,kamnitsas2017unsupervised} and retinal vessel annotation to image translation \cite{costa2017towards}. However, these methods presume targeting images have paired cross-domain data. Learning from unpaired cross-domain data is an attractive yet not well explored problem \cite{vemulapalli2015unsupervised,liu2016coupled}.

Synthesizing medical data to overcome insufficient labeled data attracted wide interests recently \cite{shrivastava2016learning,iglesias2013synthesizing,huo2017adversarial}. 
Due to the diversity of medical modalities, learning an unsupervised translation between modalities is a promising direction \cite{costa2017towards}.  
\cite{kamnitsas2017unsupervised} demonstrates the benefits on brain (MRI and CT) images, by using synthetic data as augmented training data to help lesion segmentation. 

Apart from synthesizing data, several studies \cite{kohl2017adversarial,luc2016semantic,yang2017automatic,xue2017segan} use adversarial learning as an extra supervision on the segmentation or detection networks. The adversarial loss plays a role of constraining the prediction to be close to the distribution of groundtruth. However, such strategy is a refinement process, so it is less likely to remedy the cost of data insufficiency. 


\begin{figure*}[t]
	\begin{center}
		\includegraphics[width=0.98\textwidth]{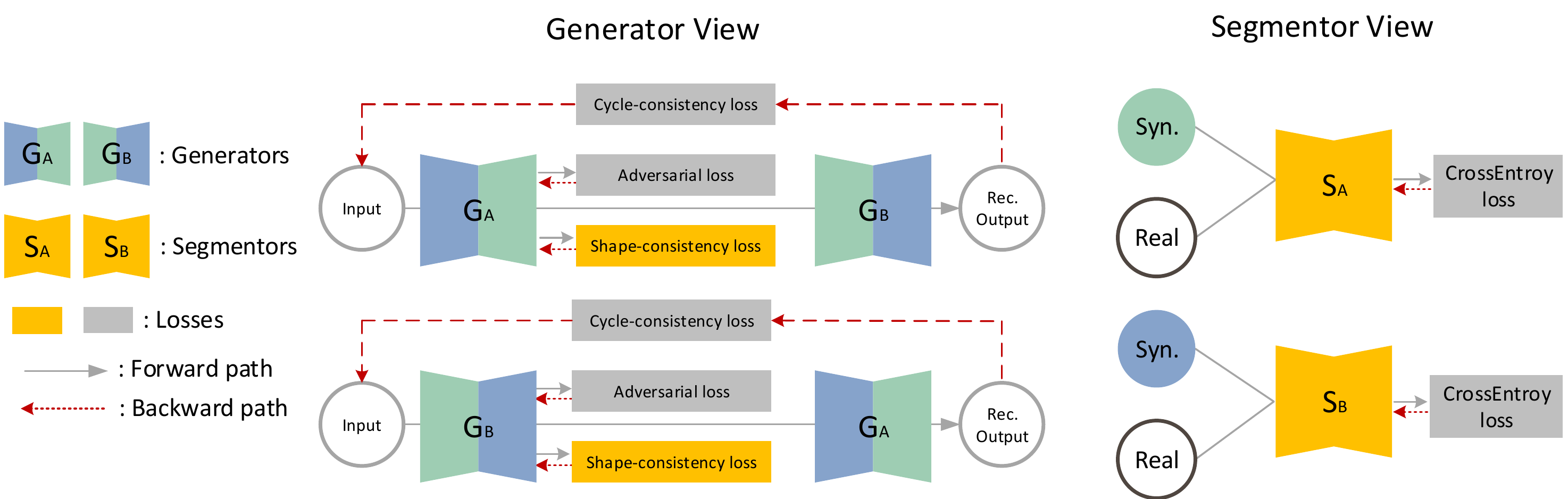}
	\end{center}
	\vspace{-.2cm}
	\caption{The illustration of our method from the generator view (left) and the segmentor view (right). \textbf{Generator view:} Two generators learn cross-domain translation between domain A and B, which are supervised by a cycle-consistency loss, a discriminative loss, and a shape-consistency loss (supported by segmentors), respectively. \textbf{Segmentor view:} Segmentors are trained by real data and extra synthetic data translated from domain-specific generators. Best viewed in color. \vspace{-.4cm}} \label{fig:arch}
\end{figure*}





\section{Proposed Method}
This section introduces our proposed method. We begin by discussing the recent advances for image-to-image translation and clarify their problems when used for medical volume-to-volume translation. Then we introduce our proposed medical volume-to-volume translation, with adversarial, cycle-consistency and shape-consistency losses, as well as dual-modality segmentation. Figure \ref{fig:arch} illustrates our method.

\subsection{Image-to-Image Translation for Unpaired Data}
GANs have been widely used for image translation in the applications that need pixel-to-pixel mapping, such as image style transfer \cite{zhang2016colorful}. ConditionalGAN \cite{isola2016image} shows a strategy to learn such translation mapping with a conditional setting to capture structure information.
However, it needs paired cross-domain images for the pixel-wise reconstruction loss. For some types of translation tasks, acquiring paired training data from two domains is difficult or even impossible. Recently, CycleGAN \cite{zhu2017unpaired} and other similar methods \cite{kim2017learning,yi2017dualgan} are proposed to generalize ConditionalGAN to address this issue.  Here we use CycleGAN to illustrate the key idea.

Given a set of unpaired data from two domains, $A$ and $B$, CycleGAN learns two mappings, $G_B: A \rightarrow B$ and  $G_A: B \rightarrow A$, with two generators $G_A$ and $G_B$, at the same time. To bypass the infeasibility of pixel-wise reconstruction with paired data, i.e. $G_B(A) \approx B$ or $G_A(B) \approx A$, CycleGAN introduces an effective cycle-consistency loss for $G_A(G_B(A)) \approx A$ and $G_B(G_A(B)) \approx B$. The idea is that the generated target domain data is able to return back to the exact data in the source domain it is generated from.
To guarantee the fidelity of fake data $G_B(A)$ and $G_A(B)$, CycleGAN uses two discriminators $D_A$ and $D_B$ to distinguish real or synthetic data and thereby encourage generators to synthesize realistic data \cite{goodfellow2014generative}. 

\subsection{Problems in Unpaired Volume-to-Volume Translation}
Lacking supervision with a direct reconstruction error between $G_B(A)$ and $B$ or $G_A(B)$ and $A$ brings some uncertainties and difficulties towards to the desired outputs for more specified tasks. And it is even more challenging when training on 3D CNNs.

To be specific, cycle-consistency has an intrinsic ambiguity with respect to geometric transformations.
For example, suppose generation functions, $G_A$ and $G_B$, are cycle consistent, e.g., $G_A(G_B(A)) = A$.
Let $T$ be a bijective geometric transformation (e.g., translation, rotation, scaling, or even nonrigid transformation) with inverse transformation $T^{-1}$.

It is easy to show that $G_A^{'} = G_A \circ T$ and $G_B^{'} = G_B \circ T^{-1}$ are also cycle consistent.
Here, $\circ$ denotes the concatenation operation of two transformations.
That means, using CycleGAN, when an image is translated from one domain to the other it can be geometrically distorted.
And the distortion can be recovered when it is translated back to the original domain without provoking any penalty in data fidelity cost.
From the discriminator perspective, geometric transformation does not change the realness of synthesized images since the shape of training data is arbitrary. 

Such problem can destroy anatomical structures in synthetic medical volumes, which, however, has not being addressed by existing methods.

\subsection{Volume-to-Volume Cycle-consistency}
To solve the task of learning generators with unpaired volumes from two domains, $A$ and $B$, we adopt the idea of the cycle-consistency loss (described above) for generators $G_A$ and $G_B$ to force the reconstructed synthetic sample $G_A(G_B(x_A))$ and $G_B(G_A(x_B))$ to be identical to their inputs $x_A$ and $x_B$:
\begin{equation}
\begin{split}
\mathcal{L}_{cyc}(G_A, G_B) & = \mathbb{E}_{x_A \sim p_{d}(x_A)}[ ||G_A(G_B(x_A)) - x_A||_1 ] \\
& + \mathbb{E}_{x_B \sim p_{d}(x_B)}[ ||G_B(G_A(x_B)) - x_B||_1 ],
\end{split}
\end{equation}
where ${x_A}$ is a sample from domain $A$ and ${x_B}$ is from domain $B$. $\mathcal{L}_{cyc}$ uses the L1 loss over all voxels, which shows better visual results than the L2 loss. 

\subsection{Volume-to-Volume Shape-consistency}
To solve the intrinsic ambiguity with respect to geometric transformations in cycle-consistency as we pointed out above, our method introduces two auxiliary mappings, defined as $S_A: A \rightarrow Y $ and $S_B: B \rightarrow Y$, to constrain the geometric invariance of synthetic data. 
They map the translated data from respective domain generators into a shared shape space $Y$ (i.e. a semantic label space) and compute pixel-wise semantic ownership. The two mappings are represented by two CNNs, namely segmentors. 
We use them as extra supervision on the generators to support shape-consistency (see Figure \ref{fig:arch}), by optimizing
\begin{equation}
\begin{split}
\mathcal{L}_{shape}(S_A, & \; S_B,G_A,G_B) = \\
& \mathbb{E}_{x_B \sim p_{d}(x_B)} [- \frac{1}{N} \sum_{i} y_B^i log(S_A(G_A(x_B))_i)] \\
+ & \mathbb{E}_{x_A \sim p_{d}(x_A)} [ - \frac{1}{N} \sum_{i} y_A^i log(S_B(G_B(x_A))_i)], \\
\end{split}
\end{equation}
where $y_A, y_B \in Y$ denote the groundtruth shape representation of sample volumes $x_A$ and $x_B$, respectively, where $y_A^i, y_B^i \in \{0,1,...,C\}$ represent one voxel with one out of $C$ classes. $N$ is the total number of voxels in a volume. $\mathcal{L}_{shape}$ is formulated as a standard multi-class cross-entropy loss. 

\vspace{.1cm}
\noindent
\textbf{Regularization} Shape-consistency provides a level of regularization on generators. Recall that different from ConditionalGAN, since we have no paired data, the only supervision for $G_A(x_B)$ and $G_B(x_A)$ is the adversarial loss, which is not sufficient to preserve all types of information in synthetic images, such as the annotation correctness. 
\cite{shrivastava2016learning} introduces a self-regularization loss between an input image and an output image to force the annotations to be preserved. Our shape-consistency performs a similar role to preserve pixel-wise semantic label ownership, as a way to regularize the generators and guarantee the anatomical structure invariance in medical volumes.

\subsection{Multi-modal Volume Segmentation}
The second parallel task we address in our method is to make use of synthetic data for improving the generalization of segmentation network, which is trained together with generators. 
From the segmentor view (Figure \ref{fig:arch}) of $S_A$ and $S_B$,  the synthetic volumes $\{G_B(x_A), y_A\}$ and $\{G_A(x_B), y_B\}$ provide extra training data to help improve the segmentors in an online manner. During training, $S_A$ and $S_B$ take both real data and synthetic data that are generated by generators online (see Figure \ref{fig:arch}). 
By maximizing the usage of synthetic data, we also use reconstructed synthetic data, $\{G_A(G_B(x_A)), y_A\}$ and $\{G_B(G_A(x_B)), y_B\}$, as the inputs of segmentors. 

Note that the most straightforward way to use synthetic data is fusing them with real data and then train a segmentation CNN. We denote this as an ad-hoc offline data augmentation approach. Compared with it, our method implicitly performs data augmentation in an online manner. Formulated in our optimization objective, our method can use synthetic data more adaptively, which thereby offers more stable training and thereby better performance than the offline approach. We will demonstrate this in experiments.

\subsection{Objective}
Given the definitions of cycle-consistency and shape-consistency losses above, we define our full objective as:  
\begin{equation}
\begin{split}
\mathcal{L}(G_{A}, G_{B},  D_{A}, D_{B}, & \; S_{A}, S_{B})  = \mathcal{L}_{GAN}(G_A, D_A) \\
& + \mathcal{L}_{GAN}(G_B, D_B) \\
& + \lambda \mathcal{L}_{cyc}(G_A, G_B) \\
& + \gamma \mathcal{L}_{shape}(S_A,  S_B, G_A, G_B) 
\end{split}
\end{equation}
The adversarial loss $\mathcal{L}_{GAN}$ (defined in \cite{zhu2017unpaired,isola2016image}) encourages local realism of synthetic data (see architecture details). $\lambda$ is set to $10$ and $\gamma$ is set to $1$ during training. 
To optimize $\mathcal{L}_{GAN}$ , $\mathcal{L}_{cyc}$, and $\mathcal{L}_{shape}$, we update them alternatively: optimizing $G_{A/B}$ with $S_{A/B}$ and $D_{A/B}$ fixed and then optimizing $S_{A/B}$ and $D_{A/B}$ (they are independent), respectively, with $G_{A/B}$ fixed. 

The generators and segmentors are mutually beneficial, because to make the full objective optimized, the generators have to generate synthetic data with lower shape-consistency loss, which, from another angle, indicates lower segmentation losses over synthetic training data.  

\begin{figure}[t]
	\begin{center}
		\includegraphics[width=0.49\textwidth]{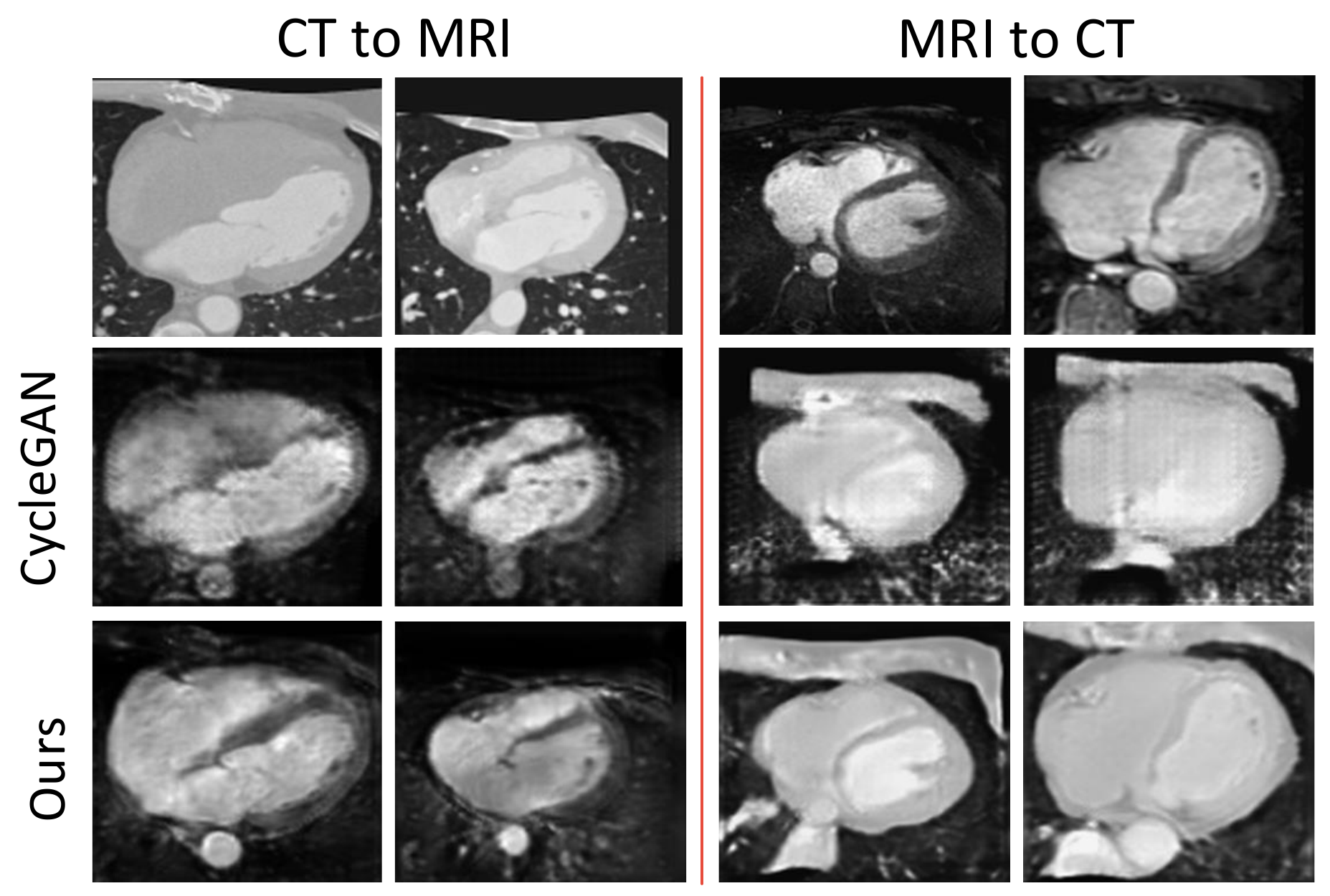}
	\end{center}
	\vspace{-.2cm}
	\caption{Example outputs on 2D slides of 3D cardiovascular CT and MRI images of the results using 3D CycleGAN (second row) and ours (third row). The first row is the input samples.
	The original results of CycleGAN have severe artifacts, checkerboard effects, and missing anatomies (e.g., descending aorta and spine), while our method overcomes these issues and achieves significantly better quality.} \label{fig:qcomare1}
\end{figure}

\begin{figure*}[t]
	\begin{center}
		\includegraphics[width=0.99\textwidth]{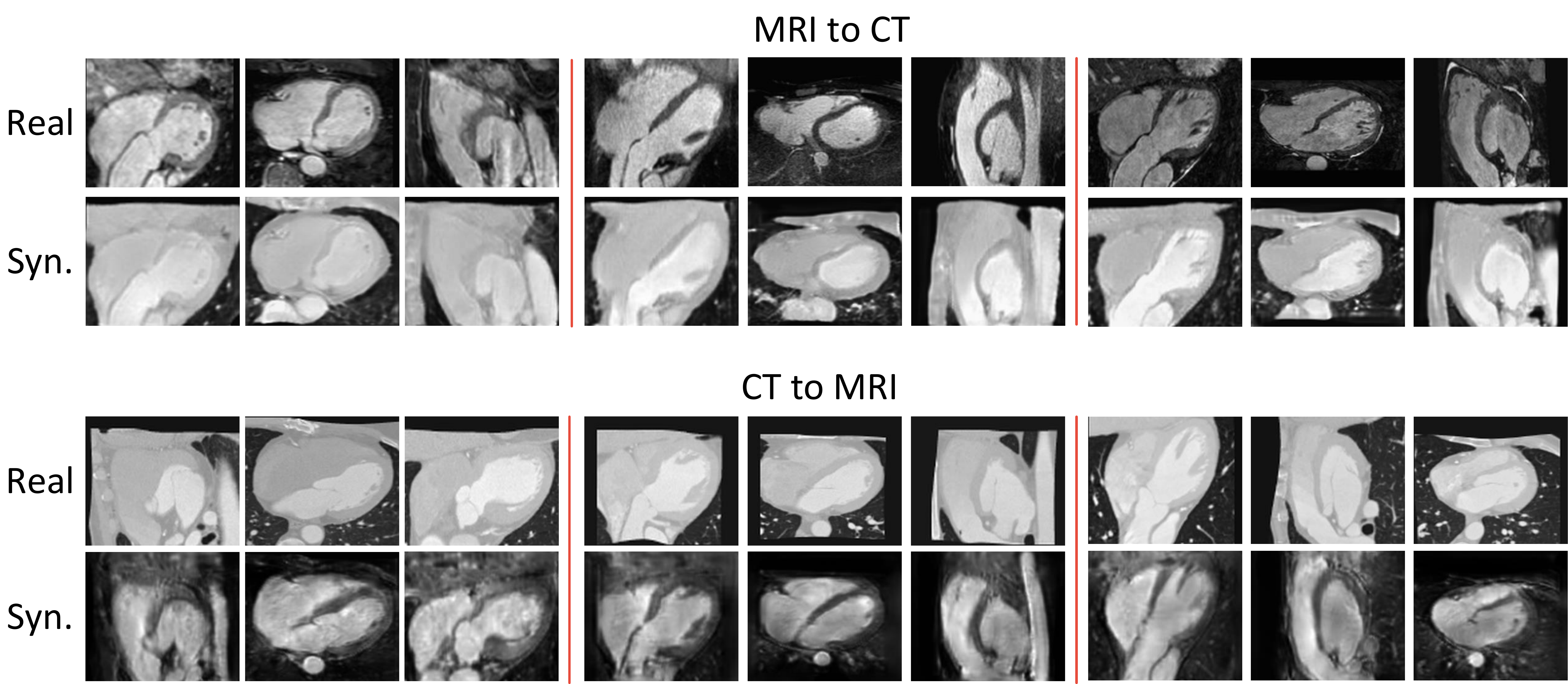}
	\end{center}
	\vspace{-.4cm}
	\caption{Qualitative results of our translation from MRI to CT (first row) and from CT to MRI (second row). For each sample (in one out of six grids), we show three orthogonal cuts through the center of 3D volumes. } \label{fig:synthetic_data} 
\end{figure*}

\section{Network Architecture and Details}
This section discusses necessary architecture and training details for generating high-quality 3D images.
\subsection{Architecture}
Training deep networks end-to-end on 3D images is much more difficult (from optimization and memory aspects) than 2D images. 
Instead of using 2.5D \cite{roth2014new} or sub-volumes \cite{kamnitsas2017unsupervised}, our method directly deals with holistic volumes. Our design trades-off network size and maximizes its effectiveness.
There are several keys of network designs in order to achieve visually better results. 
The architecture of our method is composed by 3D fully convolutional layers with instance normalization \cite{ulyanov2016instance} (performs better than batch normalization \cite{ioffe2015batch}) and ReLU for generators or LeakyReLU for discriminators. 
CycleGAN originally designs generators with multiple residual blocks \cite{he2016deep}.
Differently, in our generators, we make several critical modifications with justifications.

First, we find that using both bottom and top layer representations are critical to maintain the anatomical structures in medical images. We use long-range skip-connection in U-net \cite{ronneberger2015u} as it achieves much faster convergence and locally smooth results. ConditionalGAN also uses U-net generators, but we do not downsample feature maps as greedily as it does. We apply $3$ times downsampling with stride-2 $3{\times}3{\times}3$ convolutions totally, so the maximum downsampling rate is $8$. The upsampling part is symmetric. Two sequential convolutions are used for each resolution, as it performs better than using one.
Second, we replace transpose-convolutions to stride $2$ nearest upsampling followed by a $3{\times}3{\times}3$ convolution to realize upsampling as well as channel changes. It is also observed in \cite{odena2016deconvolution} that transpose-convolution can cause checkerboard artifacts due to the uneven overlapping of convolutional kernels. Actually, this effect is even severer for 3D transpose-convolutions as one pixel will be covered by $2^3$ overlapping kernels (results in 8 times uneven overlapping).  Figure \ref{fig:qcomare1} compares the results with CycleGAN, demonstrating that our method can obtain significantly better visual quality\footnote{We have experimented many different configurations of generators and discriminators. All trials did not achieve desired visual results compared with our configuration. }. 

For discriminators, we adopt the PatchGAN proposed by \cite{shrivastava2016learning} to classify whether an overlapping sub-volume is real or fake, rather than to classify the whole volume. Such approach limits discriminators to use unexpected information from arbitrary volume locations to make decisions. 

For segmentors, we use an U-Net \cite{ronneberger2015u}, but without any normalization layer. Totally 3 times symmetric downsampling and upsampling are performed by stride $2$ max-poling and nearest upsampling. For each resolution, we use two sequential $3{\times}3{\times}3$ convolutions.

\subsection{Training details}
We use the Adam solver \cite{kingma2014adam} for segmentors with a learning rate of $2e{-}4$ and closely follow the settings in CycleGAN to train generators with discriminators. 
In the next section, for the purpose of fast experimenting, we choose to pre-train the $G_{A/B}$ and $D_{A/B}$ separately first and then train the whole network jointly. We hypothesized that fine-tuning generators and segmentors first is supposed to have better performance because they only affect each other after they have the sense of reasonable outputs.
Nevertheless, we observed that training all from scratch can also obtain similar results. 
It demonstrates the effectiveness to couple both tasks in an end-to-end network and make them converge harmonically.
We pre-train segmentors for $100$ epochs and generators for $60$ epochs. After jointly training for $50$ epochs, we decrease the learning rates for both generators and segmentors steadily for $50$ epochs till 0. We found that if the learning rate decreases to a certain small value, the synthetic images turn to show clear artifacts and the segmentors tend to overfit. We apply early stop when the segmentation loss no longer decreases for about $5$ epochs (usually takes $40$ epochs to reach a desired point). In training, the number of training data in two domains can be different. We go through all data in the domain with larger amount as one epoch.

\begin{table}[t] 
	\caption{Shape quality evaluation using the proposed S-score (see text for definition) for synthesized images. The synthetic volumes using our method has much better shape quality on both modalities.  SC denotes shape-consistency.} \label{table:score}
	\vspace{-.2cm} 
	\begin{center}
		\begin{tabularx}{.29\textwidth}{c|cc}
			\specialrule{1.5pt}{0pt}{0pt}  
			\multirow{2}{*}{Method}	& \multicolumn{2}{c}{S-score ($\%$)}	\\	\cline{2-3}
			&	 CT		&	MRI		     \\ \hline
			$G$ w/o SC  &	66.8			&	67.5	\\ \hline
			$G$ w/ SC (Ours)	&	\textbf{69.2}			&	\textbf{69.6}	\\\hline
			
		\end{tabularx} \vspace{-.4cm}
	\end{center}
\end{table}

\section{Experimental Results}
This section evaluates and discusses our method. We introduce a 3D cardiovascular image dataset. Heart is a perfect example of the difficulty in getting paired cross-modality data as it is a nonrigid organ and it keeps beating. Even if there are CT and MRI scans from the same patient, they cannot be perfectly aligned. 
Then we evaluate the two tasks we addressed in our method, i.e., volume segmentation and synthesis, both qualitatively and quantitatively with our proposed auxiliary evaluation metrics.

\subsection{Dataset}
We collected 4,354 contrasted cardiac CT scans from patients with various cardiovascular diseases ($2{-}3$ volumes per patients).
The resolution inside an axial slice is isotropic and varies from 0.28 mm to 0.74 mm for different volumes.
The slice thickness (distance between neighboring slices) is larger than the in-slice resolution and varies from 0.4 mm to 2.0 mm.
In addition, we collected 142 cardiac MRI scans with a new compressed sensing scanning protocol.  
The MRI volumes have a near isotropic resolution ranging from 0.75 to 2.0 mm. This true 3D MRI scan with isotropic voxel size is a new imaging modality, only available in handful top hospitals.
All volumes are resampled to 1.5 mm for the following experiments.
We crop $86{\times}112{\times}112$ volumes around the heart center.
The endocardium of all four cardiac chambers is annotated. The left ventricle epicardium is annotated too, resulting in five anatomical regions.

We denote CT as domain $A$ data and MRI as domain $B$.  We organize the dataset in two sets $\mathcal{S}_1$ and $\mathcal{S}_2$. For $\mathcal{S}_1$, we randomly select 142 CT volumes from all CT images to match the number of MRI volumes. For both modalities, $50\%$ data is used as training and validation and the rest $50 \%$ as testing data. 
For $\mathcal{S}_2$, we use all the rest 4,212 CT volumes as an extra augmentation dataset, which is used to generate synthetic MRI volumes for segmentation.
We fix the testing data in $\mathcal{S}_1$ for all experiments.

\subsection{Cross-domain Translation Evaluation}
We evaluate the generators both qualitatively and quantitatively.
Figure \ref{fig:synthetic_data} shows some typical synthetic results of our method.
As can be observed visually, the synthetic images are close to real images and no obvious geometric distortion is introduced during image translation. Our method well preserves cardiac anatomies like aorta and spine. 
                                     
\noindent
\textbf{Shape invariance evaluation}  For methods of GANs to generate class-specific natural images, \cite{salimans2016improved} proposes to use the Inception score to evaluate the diversity of generated images, by using an auxiliary trained classification network.

Inspired by this, we propose the S-core (segmentation score) to evaluate the shape invariance quality of synthetic images. We train two segmentation networks on the training data of respective modalities and compare the multi-class Dice score of synthetic volumes. For each synthetic volume, S-score is computed by comparing to the groundtruth of the corresponding real volume it is translated from. Hence, higher score indicates better matched shape (i.e. less geometric distortion). 
Table \ref{table:score} shows the S-score of synthetic data from CT and MRI for generators without the shape-consistency loss, denoted as $G$ w/o SC. Note that it is mostly similar with CycleGAN but using our optimized network designs. As can be seen, our method ($G$ w/ SC) with shape-consistency achieves large improvement over the baseline on both modalities. 

\begin{figure}[t]
	\begin{center}
		\includegraphics[width=0.45\textwidth]{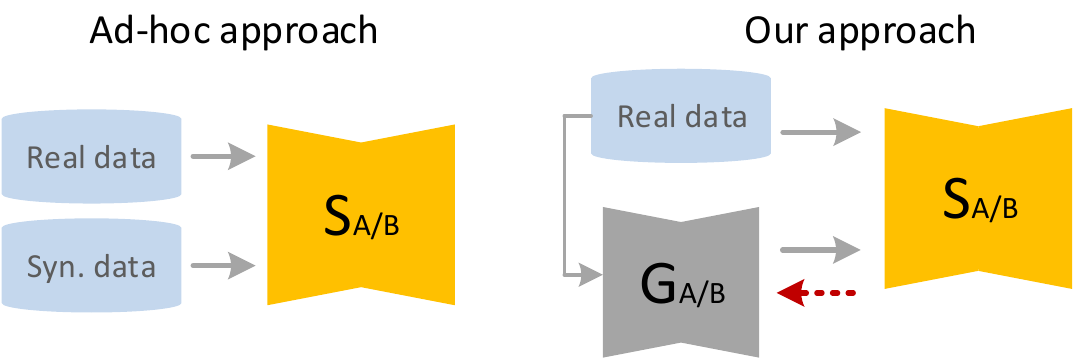}
	\end{center}
	\vspace{-.2cm}
	\caption{Illustration of the strategies to use synthetic data to improve segmentation. The left is the comparing ad-hoc offline approach. The right in our approach that uses synthetic data from the generator in an online fashion.} \label{fig:onoffline}
\end{figure}

\begin{table}[t] 
	\caption{The segmentation performance comparison. Initialized from the baseline model trained with only \textbf{R}eal data (Baseline (R)), the second and third rows show the boosted results by using \textbf{S}ynthetic data with the comparing ADA and our method, respectively.  }
	\vspace{-.4cm}
	\label{table:segmentation} 
	\begin{center}
		\begin{tabularx}{.27\textwidth}{c|cc}
			\specialrule{1.5pt}{0pt}{0pt}  
			\multirow{2}{*}{Method}	& \multicolumn{2}{c}{Dice score ($\%$)}	\\	\cline{2-3}
			&	 CT	&	MRI		     \\ \hline
			Baseline (R) &		67.8			&	70.3		\\ \hline
			ADA (R+S) &	66.0			&	71.0	\\ \hline
			Ours (R+S)	&	\textbf{74.4}	&	\textbf{73.2}	\\ \hline
		\end{tabularx} \vspace{-.4cm}
	\end{center}
\end{table}

\subsection{Segmentation Evaluation} 
Here we show how well our method can use the synthetic data and help improve segmentation. We compare to an ad-hoc approach as we mentioned above.  
Specifically, we individually train two segmentors, denoted as $\tilde{S_A}$ and $\tilde{S_B}$. We treat the segmentation performance of them as Baseline (R) in the following. Then we train generators $\tilde{G_A}$ and $\tilde{G_B}$  with the adversarial and cycle-consistency losses (setting the weight of the shape-consistency loss to 0). Then by adding synthetic data, we perform the following comparison:

\begin{enumerate}[topsep=0pt]
	\item Ad-hoc approach (ADA): We use $\tilde{G_A}$ and $\tilde{G_B}$ to generate synthetic data (To make fair comparison, both synthetic data  $G_{A/B}(x_{B/A})$ and reconstructed data $G_{A/B}(G_{B/A}(x_{A/B}))$ are used). We fine-tune $\tilde{S}_{A/B}$ using synthetic together with real data (Figure \ref{fig:onoffline} left)\footnote{At each training batch, we take half real and half synthetic data to prevent possible distraction from low-quality synthetic data.}. 
	\item Our method: We join $\tilde{S_A}$,  $\tilde{S_B}$, $\tilde{G_A}$, and $\tilde{G_B}$ (also with discriminators) and fine-tune the overall networks in an end-to-end fashion (Figure \ref{fig:onoffline} right), as specified in the training details. 
\end{enumerate}
Note that the comparing segmentation network is U-net \cite{ronneberger2015u}. For medical image segmentation, U-Net is well recognized as one of the best end-to-end CNN. Its long-range skip connection performs usually better or equal well as FCN or ResNet/DenseNet based architectures do \cite{drozdzal2016importance}, especially for small size medical datasets. The results of U-net is very representative for state-of-the-art medical image segmentation on our dataset. 

We perform this experimental procedure on $\mathcal{S}_1$ and $\mathcal{S}_2$ both.
In the the first experiment on $\mathcal{S}_1$,
we test the scenario that how well our method uses synthetic data to improve segmentation given only limited real data. Since we need to vary the number of data in one modality and fix another, we perform the experiments on both modalities, respectively. 

By using $14\%$  real data and all synthetic data from the counter modality, Table \ref{table:segmentation} compares the segmentation results. We use the standard multi-class Dice score as the evaluation metric \cite{dice1945measures}.
As can be observed, our method achieves much better performance on both modalities. For CT segmentation, ADA even deteriorates the performance.  We speculate that it is because the baseline model trained with very few real data has not been stabilized. Synthetic data distracts optimization when used for training offline. While our method adapts them fairly well and leads to significant improvement.

\begin{figure}[t]
	\begin{center}
		\includegraphics[width=0.48\textwidth]{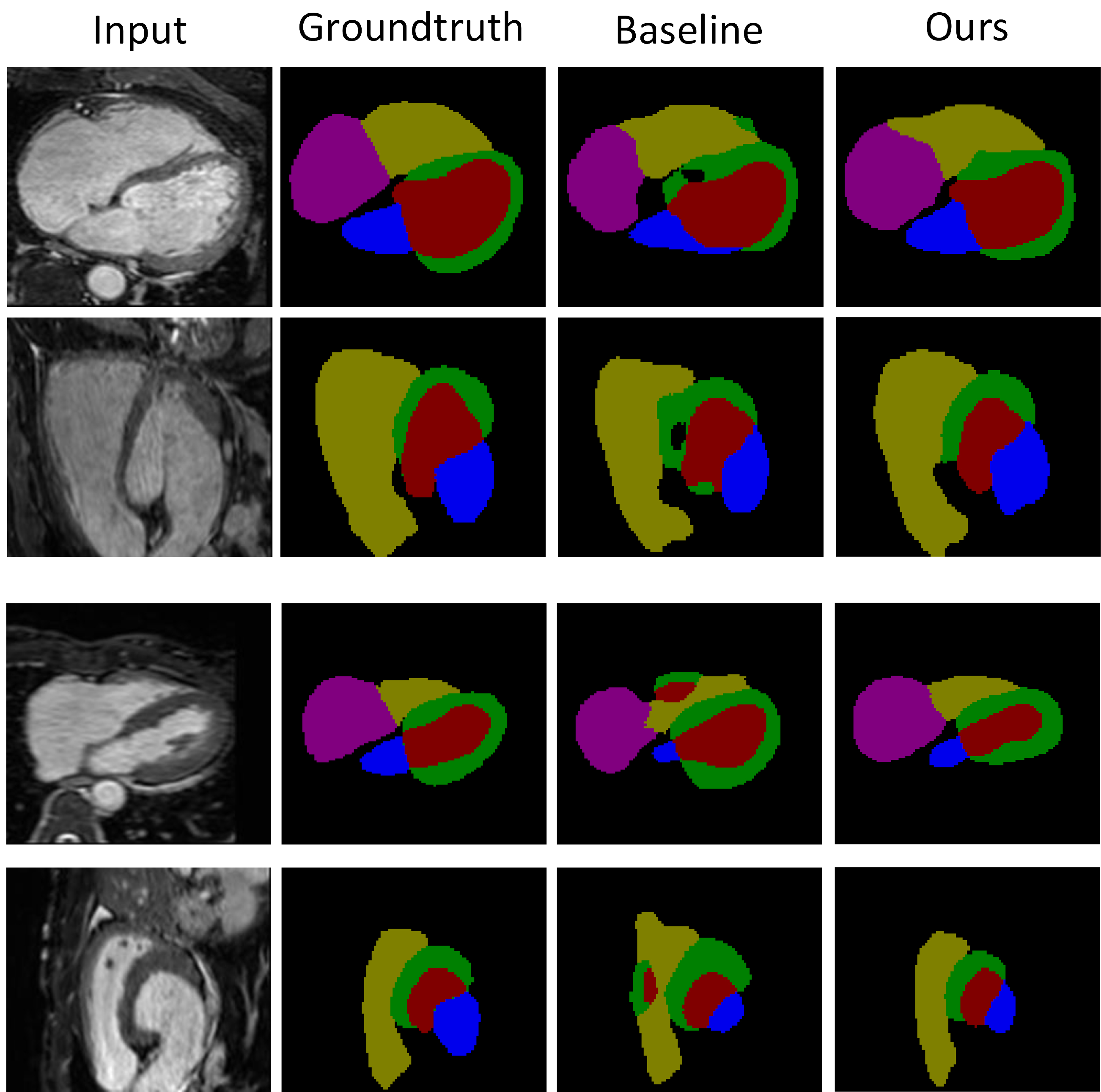}
	\end{center}
	\vspace{-.2cm}
	\caption{The qualitative evaluation of segmentation results on MRI. We show the axial and sagittal views of two samples. Our method boosts the baseline segmentation network with only extra synthetic data. As can be observed, the segmentation errors of the baseline are largely corrected. } \label{fig:segmentation}
\end{figure}

\begin{figure*}[t]	
	\begin{center}
		\begin{subfigure}[b]{0.33\textwidth}
			\centering
			\includegraphics[width=0.999\textwidth, height=0.6\textwidth]{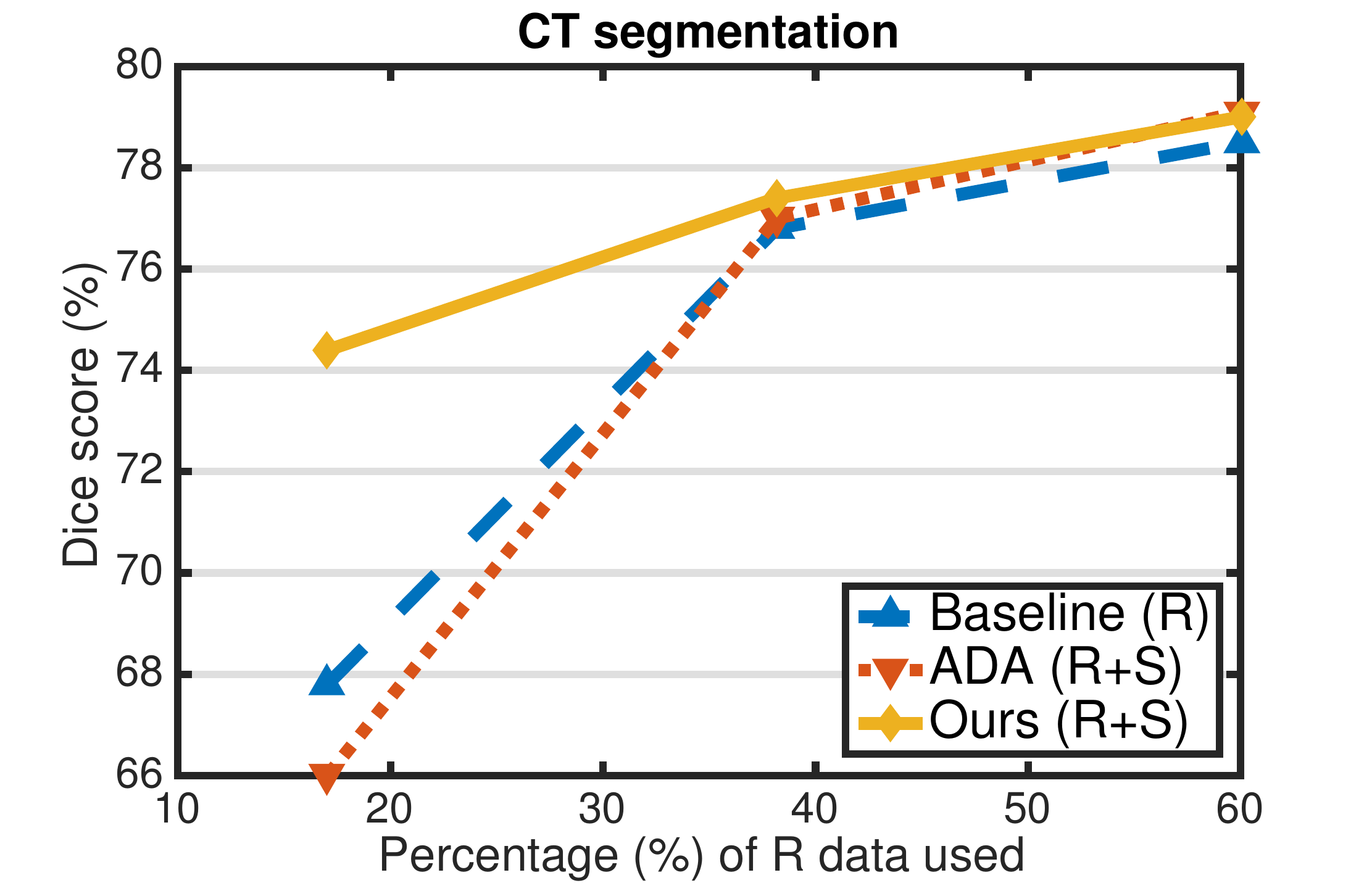}
		\end{subfigure}
		\begin{subfigure}[b]{0.33\textwidth}
			\centering
			\includegraphics[width=.999\textwidth, height=0.6\textwidth]{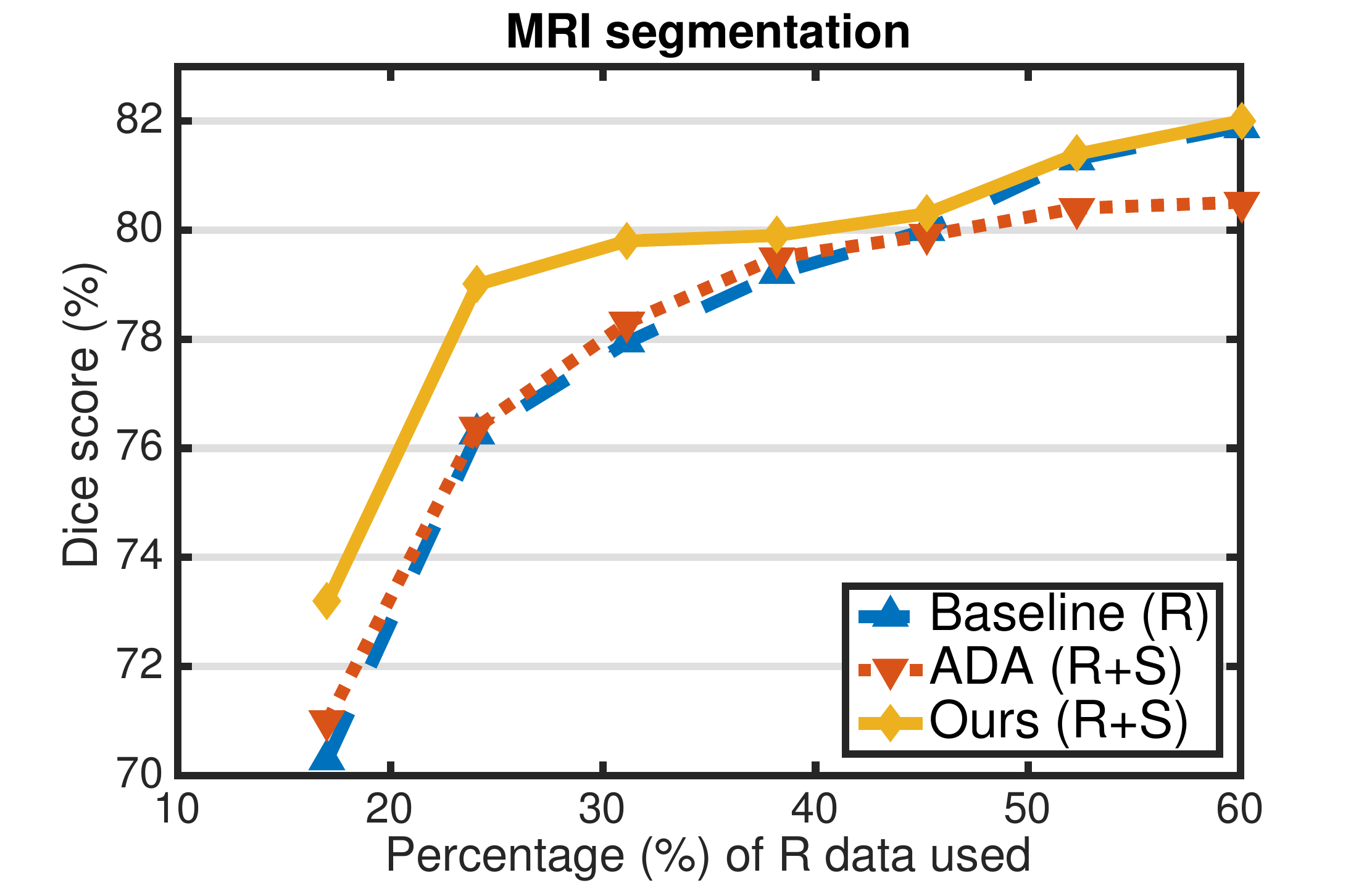}
		\end{subfigure}
		\begin{subfigure}[b]{0.33\textwidth}
			\centering
			\includegraphics[width=.999\textwidth, height=0.6\textwidth]{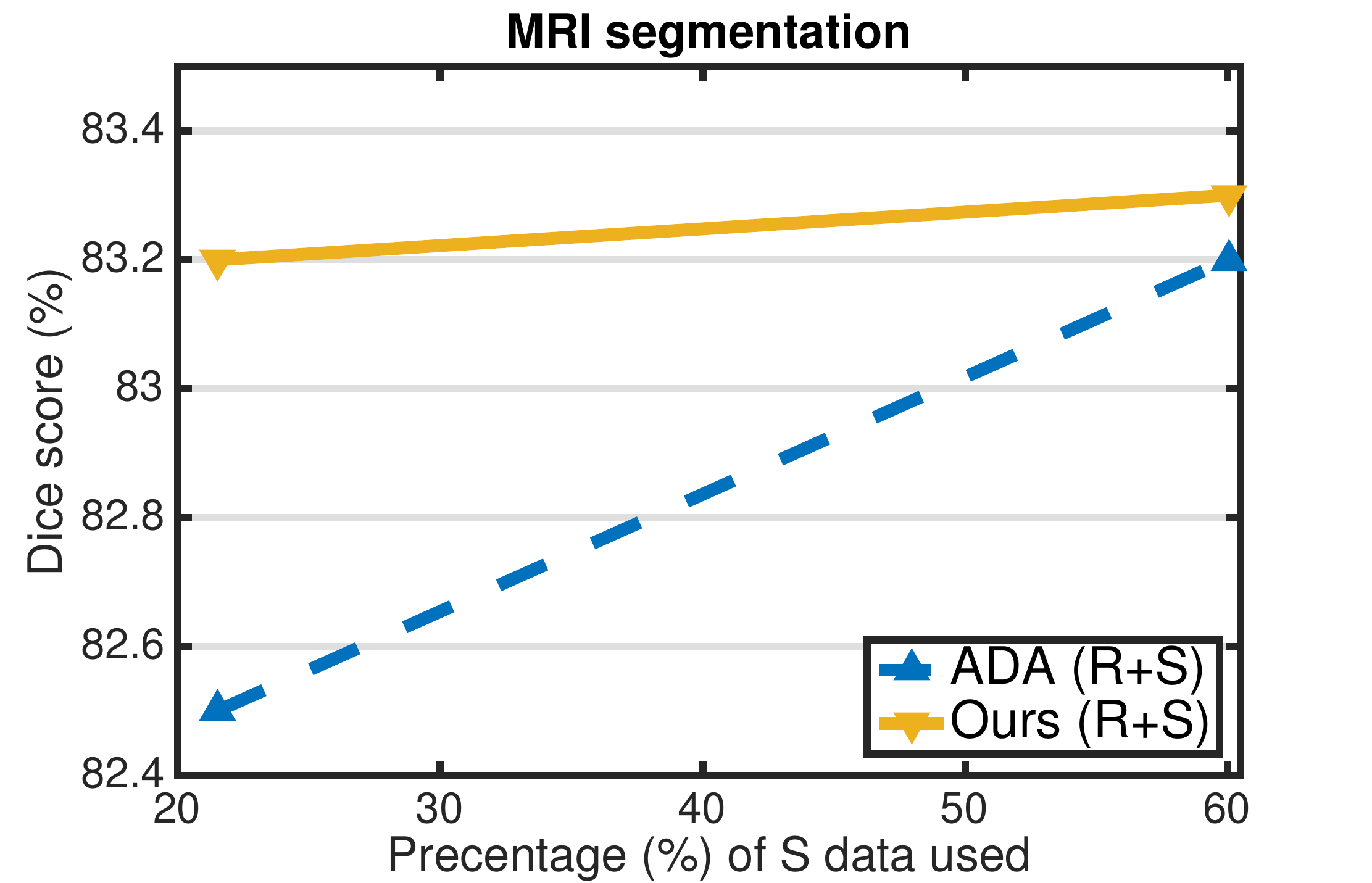}
		\end{subfigure}
		
	\end{center}
	\vspace{-.4cm}
	\caption{The segmentation accuracy (mean Dice score) comparison to demonstrate the effectiveness of our method of using \textbf{S}ynthetic data to boost segmentation. The left plot shows the segmentation accuracy by varying the percentage of \textbf{R}eal data used for training segmentation on CT using dataset $\mathbf{S}_1$, using a equal number of synthetic data. Baseline (R) is trained with only real data. Others are trained from it, e.g. ADA (R+S) is trained by adding only \textbf{S} data. The middle plot shows the same experiments on MRI. The right plot shows results by varying the number of synthetic data on MRI using dataset $\mathbf{S}_2$ using a equal number of real data. Our method has consistently better performance. See text for details about comparing methods. \vspace{-.4cm}} \label{fig:seg_eval} 
\end{figure*}

We also demonstrate the qualitative results of our method in Figure \ref{fig:segmentation}. By only using extra synthetic data, our method largely corrects the segmentation errors.
Furthermore, we show the results by varying the number of real data used in Figure \ref{fig:seg_eval} (left and middle). Our method has consistently better performance than the ADA. In addition, we notice the increment is growing slower as the number of real data increases. One reason is that more real data makes the segmentors get closer to its capacity, so the effect of extra synthetic data gets smaller. But this situation can be definitely balanced out by increasing the size of segmentors with sufficient GPU memory.


\begin{figure}[t]
	\begin{center}
		\includegraphics[width=0.47\textwidth]{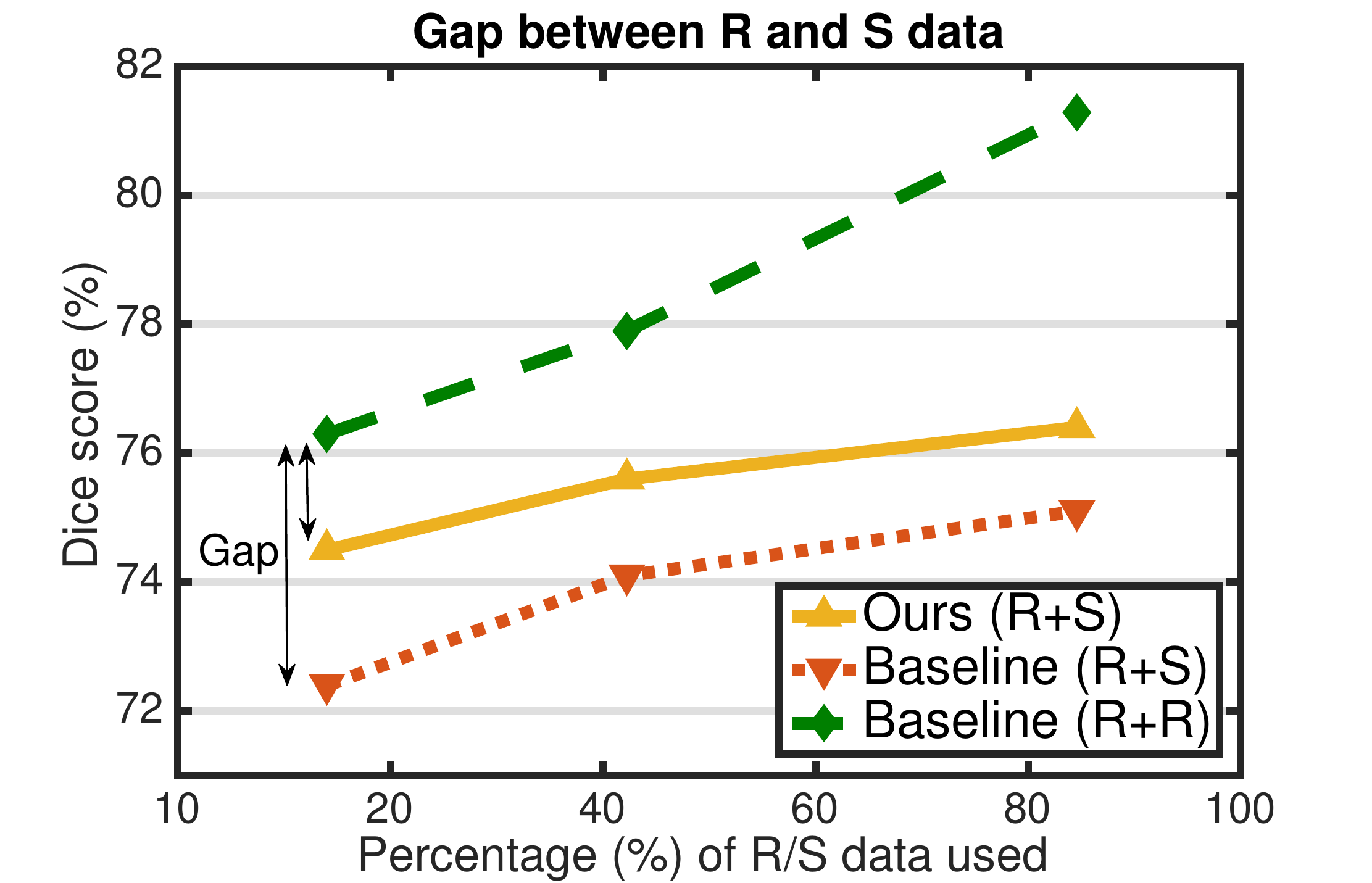}
	\end{center}
	\vspace{-.4cm}
	\caption{The gap analysis of \textbf{R}eal and \textbf{S}ynthetic data. For all comparing methods, we use one pre-trained network with $14\%$ real data, whose Dice score is $70.3\%$. Then we vary the number of R or S data used to boost segmentation of Baseline (R+R),  Baseline (R+S), and Ours (R+S).  Our method significantly reduces the gap for all settings.} \label{fig:gap} 
\end{figure}

The second experiment is applied on $\mathcal{S}_2$, which has much more CT data, so we aim at boosting the MRI segmentor. We vary the number of used synthetic data and use all real MRI data. Figure \ref{fig:seg_eval} (right) compares the results.  Our method still shows better performance. As can be observed, our method uses $23\%$ synthetic data to reach the accuracy of the ADA when it uses $100\%$ synthetic data.

\vspace{-.2cm}
\subsection{Gap between synthetic and real data}
Reducing the distribution gap between real and synthetic data is the key to make synthetic data useful for segmentation. 
Here we show a way to interpret the gap between synthetic and real data by evaluating their performance to improve segmentation. On dataset $\mathcal{S}_1$,
we train a MRI segmentor using $14\%$ real data. Then we boost the segmentor by adding 1) pure MRI real data, 2) using ADA, and 3) using our method. 
As shown in Figure \ref{fig:gap}, our method reduces the gap of the ADA significantly, i.e., by $61\%$ given $14\%$ real data and $20.9\%$ given $85\%$ real data.

Moreover, we found that, when using the synthetic data as augmented data offline (our comparing baseline), too much synthetic data could diverge the network training. While in our method, we did not observe such situation. However, we also observe that the gap is more difficult to reduce as the number of read data increases. Although one of reasons is due to the modal capacity, we believe the solution of this \textit{gap-reduction} worth further study.

\vspace{-.1cm}
\section{Conclusion}
In this paper, we present a method that can simultaneously learn to translate and segment medical 3D images, which are two significant tasks in medical imaging. Training generators for cross-domain volume-to-volume translation is more difficult than that on 2D images. We address three key problems that are important in synthesizing realistic 3D medical images: 1) learn from unpaired data, 2) keep anatomy (i.e. shape) consistency, and 3) use synthetic data to improve volume segmentation effectively. We demonstrate that our unified method that couples the two tasks is more effective than solving them exclusively. Extensive experiments on a 3D cardiovascular dataset validate the effectiveness and superiority of our method.

{\small
\bibliographystyle{ieee}
\bibliography{egbib}
}

\end{document}